\documentclass{article}
\usepackage[utf8]{inputenc}

\usepackage{amsmath}
\usepackage{amssymb}
\usepackage{mathcomp}
\usepackage{graphicx}
\usepackage{siunitx}
\usepackage{float}
\usepackage{listings}
\usepackage{subcaption}
\usepackage[export]{adjustbox}
\usepackage{siunitx} % Required for alignment

\usepackage{xcolor}

\definecolor{codegreen}{rgb}{0,0.6,0}
\definecolor{codegray}{rgb}{0.5,0.5,0.5}
\definecolor{codepurple}{rgb}{0.58,0,0.82}
\definecolor{backcolour}{rgb}{0.95,0.95,0.92}

\lstdefinestyle{mystyle}{
    backgroundcolor=\color{backcolour},   
    commentstyle=\color{codegreen},
    keywordstyle=\color{magenta},
    numberstyle=\tiny\color{codegray},
    stringstyle=\color{codepurple},
    basicstyle=\ttfamily\footnotesize,
    breakatwhitespace=false,         
    breaklines=true,                 
    captionpos=b,                    
    keepspaces=true,                 
    showspaces=false,                
    showstringspaces=false,
    showtabs=false,                  
    tabsize=2
}

\usepackage{url}
\graphicspath{ {./figures/} }
\usepackage{biblatex}
\title{Fast 3D Image Moments}
\author{William Diggin
     \and 
     Michael Diggin}
\date{December 2020}

\begin{document}

\maketitle

\begin{abstract}
An algorithm to efficiently compute the moments of volumetric images is disclosed. The approach demonstrates a reduction in processing time by reducing the computational complexity significantly. Specifically, the algorithm reduces multiplicative complexity from $\mathcal{O}(n^{3})$ to $\mathcal{O}(n)$. Several 2D projection images of the 3D volume are generated. The algorithm computes a set of 2D moments from those 2D images. Those 2D moments are then used to derive the 3D volumetric moments. Examples of use in MRI or CT and related analysis demonstrates the benefit of the Discrete Projection Moment Algorithm. The approach is also useful in computing the moments of a 3D object using a small set of 2D tomographic images of that object.
\end{abstract}

\section{Introduction}

Two dimensional image moments were first defined by Hu \cite{HuArticle}. These raw moments are used to derive 7 moment invariants. These invariants are independent of position, scale and orientation. They are useful as feature descriptors for objects in images. For a 2-D image $I(i,j)$ of size $M\!\cdot\!N$, where $0\leq i<M, 0\leq j<N$ the discrete moment generating equation of order $(p\!+\!q)$ is given by: 
\begin{equation}
\label{base_moments}
    M_{pq} = \sum_{i, j} I(i,j) i^p j^q 
\end{equation}

Three dimensional moment invariants were proposed by Sadjadi and Hall \cite{SadjadiArticle} that extend Hu's 2D geometric moments to support invariant object recognition for solid homogeneous objects and objects of varying density. For a volume $V(x,y,z)$ of size $L\!\cdot\!M\!\cdot\!N$, where $0\leq x<L, 0\leq y<M, 0\leq z<N$, the discrete moment generating equation of order $(p\!+\!q\!+\!r)$ is given by: 
\begin{equation}
\label{3d_moments}
    \mathcal{M}_{pqr} = \sum_{x,y,z} V(x,y,z) x^p y^q z^r
\end{equation}

Flusser et al \cite{flusser_tensor} used a tensor and graph method to derive 1185 3D moment invariants up to order 16. These moments are invariant to translation, rotation and scaling. The research demonstrated a volume and surface based approach to their computation.

The medical domain has been an early adopter of 3D image processing, based on tomography and related areas. It is therefore not surprising that much of the research and development of using 3D moments has radiology and anatomy based applications. Reeves et al \cite{patent:7274810} used 3D densitometric moments to assist with pulmonary nodule measurement (position, orientations, size, etc) and analysis in tomography-based applications. Moments and moment-based statistics were used to measure the changes in related artifacts for the purposes of early diagnosis and intervention.

A method to use geometric moment invariants to assist in segmenting brain structures in MRI sequences provides benefits for medical image analysis and the diagnosis of neurological conditions or diseases \cite{brain_segments}.
An important research topic is registration of anatomical images, intra- or inter- subject and the use of 3D moments to provide attribute vectors that aid in the correspondence of anatomical structures \cite{hammer}. Others have used 3D moment invariants to assist with  morphometry of brain anatomical entities, e.g. cortical sulci \cite{brain_morohometry}, allowing characterization of handedness, etc. In another example, 3D moments up to $6^{th}$ order were used to generate invariant feature descriptors and were applied to fMRI data to characterise activations in different neural regions of interest \cite{fmri_moments}. Other researchers have used 3D moments to detect and track vascular networks in MRA \cite{blood_moments} or to first segment and then track arteries in CTA \cite{artery_moments} - typically using lower order moments for geometric measurements of position, scale and orientation.
Outside the medical domain, 3D moments have been used to generate geometric feature vectors that provide indexing into a CAD library or database of parts \cite{cadmoments}.

In section \ref{comparative}, we review the state of the art for 3D moment computation. In section \ref{thealgorithm}, we outline the Discrete Projection Moment Algorithm to compute the 3D moments efficiently. Then an MRI example is used to showcase the approach. In section \ref{implementation}, some of the practical implementation details are discussed followed by a presentation of comparative results that demonstrate the efficacy of the algorithm relative to the direct naïve approach and with the OpenCV Library \cite{opencv_library}. Finally, in section \ref{analysis}, we conduct some analysis of the results and capture some of the learnings.

\section{Comparative Computation Methods} \label{comparative}
Several researchers have proposed novel methods for the fast and efficient computation of moments for volumetric images. These can be considered as surface-based, composition-based or hardware-based. The surface-based approaches generally treat the 3D object as binary and uses variations of Gauss's Theorem to compute the moments from the boundaries of the object(s). The composition-based approaches operate in regions and while suitable for binary images, could also consider voxel intensities as part of the moment computation. Hardware-based approaches seek to design dedicated processing units and associated algorithms for the purposes of rapid computation. The focus of this review is to consider 1-2 leading examples in each approach.

\subsection{ Surface-based}
Yang et al \cite{Yang} \cite{yang2} developed a method using a existing surface tracking algorithm along with a discrete implementation of Gauss' Theorem. On binary volumes, they claim that their algorithm reduces the computational complexity from $\mathcal{O}(n^{3})$ to $\mathcal{O}(n^{2})$ - although not including the computational cost of the surface tracking method.

\subsection{Composition-based}
Sossa-Azuela et al \cite{Sossaarticle} use a morphological operation approach to generate a partition. The partition is a set of cubes that represent regions in the volume. The moments of the volume are simple computations of the moments of the cubes. The researchers claim that the computational complexity is $\mathcal{O}(n)$. This is true where the volume is already pre-processed into cubes/regions. However that partitioning step requires a brute-force computation. The cubes/regions are assumed to be either binary or contain the same voxel value, and therefore this algorithm is constrained by those limitations.

Li and Ma \cite{Li_Ma} presented a method that uses linear transforms on the 3D image. The moments of the linear transforms are computed and the result transformed to conventional moments. While the method demonstrated a reduction in computation, it is limited to 3D binary images.

The OpenCV Library \cite{opencv_library} has optimised algorithms for 2D image moment computation. Extending these algorithms to the 3D case not only provides a composition-based solution, but also a non-binary computational approach. The extended algorithm for 3D will be analysed in this paper.

\subsection{Hardware-based}
Berj{\'o}n et al. \cite{Berjon} developed a GPU based approach to compute moments of binary volumetric images. Employing pre-computed lookup tables, they reported a 30x improvement over a CPU based approach in many cases. Liu et al. \cite{systolic_moments} designed a systolic array for the computaion of 3D moments. Using a 3-stage system, this is able to compute the raw moments for binary and grey-level images as well as their invariant moments.

\section{Discrete Projection Moment (DPM) Algorithm} \label{thealgorithm}
Given a volumetric image of size $L\!\cdot\! M\!\cdot\! N$, to compute its moments up to the $(p\!+\!q\!+\!r)^{th}$ order $k$ using the direct, naïve method in equation \eqref{3d_moments} requires $\mathcal{O}(k^{3}\!\cdot\! LMN)$ computations (both additions and multiplications).

However, where the volumetric image is projected onto several 2D integral images, the 2D moments of the integrals can be used to compute the 3D moments of the original volume. Let $\mathcal{P}^{}$ be a integral projection of the volume function $\mathcal{V}(x,y,z)$ onto some plane $(x^{'},y^{'})$
\[
    \mathcal{P}(x^{'},y^{'},\theta,\phi)=\int_{z^{'}} \mathcal{V}(x,y,z)dz^{'}
\]
\begin{equation}
\label{3d_radon}
    where \begin{pmatrix} x^{'}\\ y^{'} \\z^{'}\end{pmatrix}\!=\!\begin{pmatrix}
    \cos\phi\cos\theta &\sin\phi\cos\theta &\sin\theta\\ \!-\!\sin\phi &\cos\phi &0\\ \!-\!\cos\phi\sin\theta &\!-\!\sin\phi\sin\theta &\cos\theta\\ \end{pmatrix}\!
    \begin{pmatrix} x\\ y\\ z\end{pmatrix}
\end{equation}

The transformation represents a rotation by an angle $\phi$ around the $z$-axis, followed by a rotation by an angle $\theta$ around the $y^{'}$-axis. The plane $(x^{'},y^{'})$ is oriented by $(\theta,\phi)$ and the volume function $\mathcal{V}(x,y,z)$ is integrated along the $z^{'}$-axis normal to that plane. The set of these projections over $0\leq\theta<\pi$ and $0\leq\phi<\pi$ is the 3D Radon Transform \cite{deans2013radon}.

\subsection{Discrete Projections}

We are interested in a small subset of projections in the discrete case. To avoid the computation cost of interpolation, we choose the angles $(\theta,\phi)$ so that the integration aligns with the voxels. The result is a discrete 2D image $I(i,j,\theta,\phi)$ that is a integer summation of its 3D discrete counterpart $V(x,y,z)$ along an axis normal to the 2D image plane:
\begin{equation} \label{Iij}
    I(i,j,\theta,\phi) = \sum_{\begin{smallmatrix}i=f(x,y,z)\\j=g(x,y,z)\end{smallmatrix}} V(x,y,z)
\end{equation}
where $f()$ and $g()$ map from 3D volume space to 2D image space.

\begin{figure}[ht]
\begin{subfigure}{0.49\textwidth}
\includegraphics[scale=0.32]{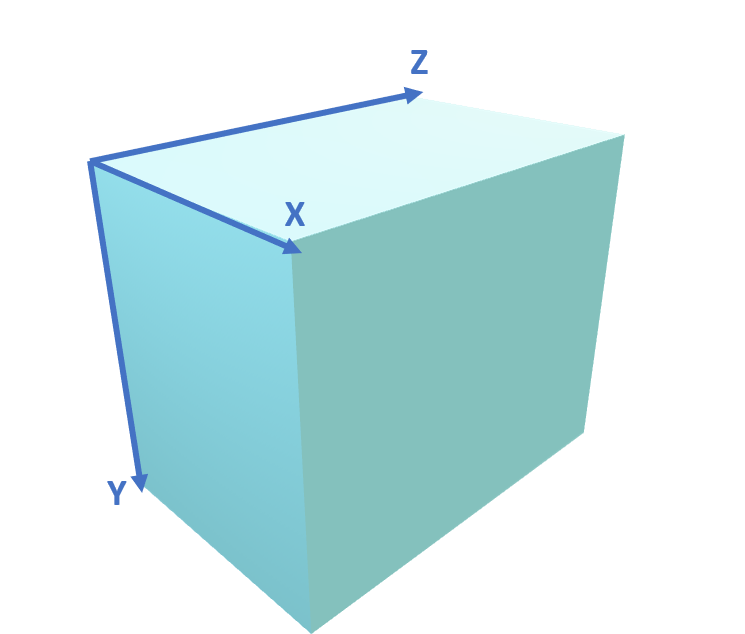}
\caption{$V(x,y,z)$ of size $L\!\cdot\! M\!\cdot\! N$}
\label{fig:base}
\end{subfigure}
\begin{subfigure}{0.5\textwidth}
\includegraphics[scale=0.4]{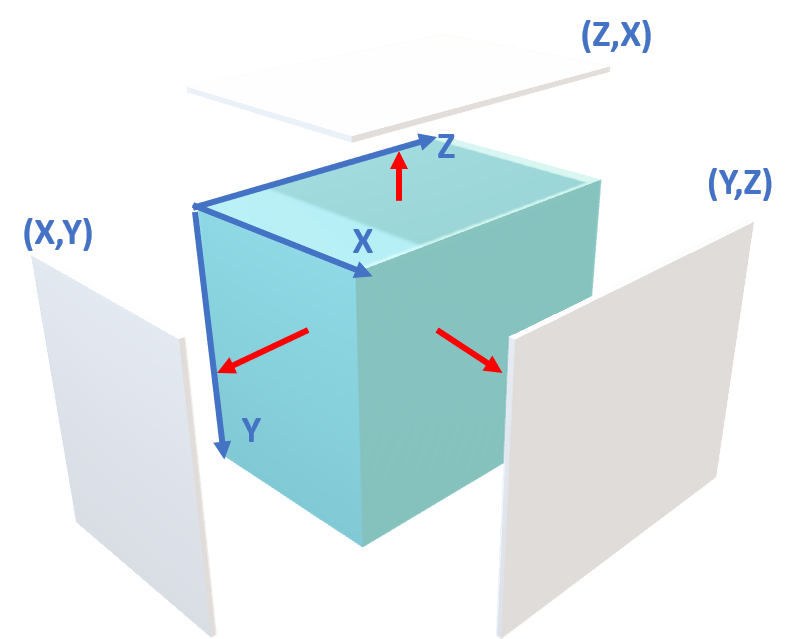}
\caption{Projections along each axis}
\label{fig:0projects}
\end{subfigure}

\begin{subfigure}{0.5\textwidth}
\includegraphics[scale=0.4, left]{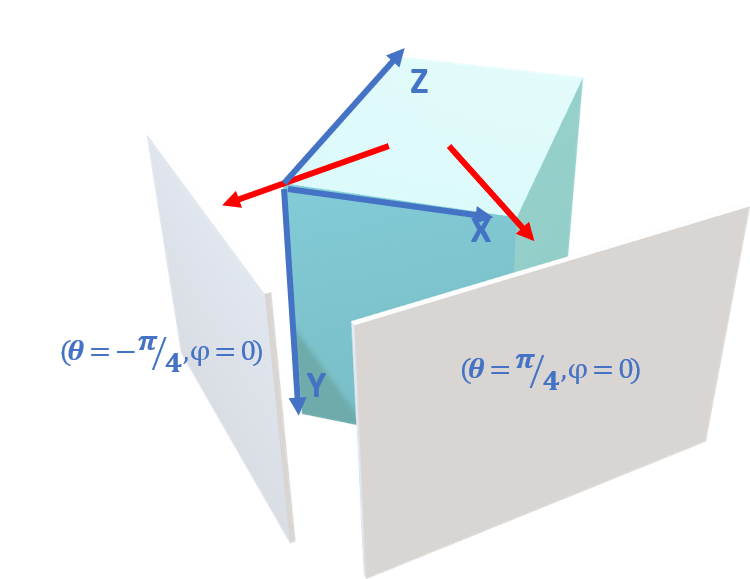}
\caption{Example axonometric projections}
\label{fig:45projects}
\end{subfigure}
\begin{subfigure}{0.49\textwidth}
\includegraphics[scale=0.4, right]{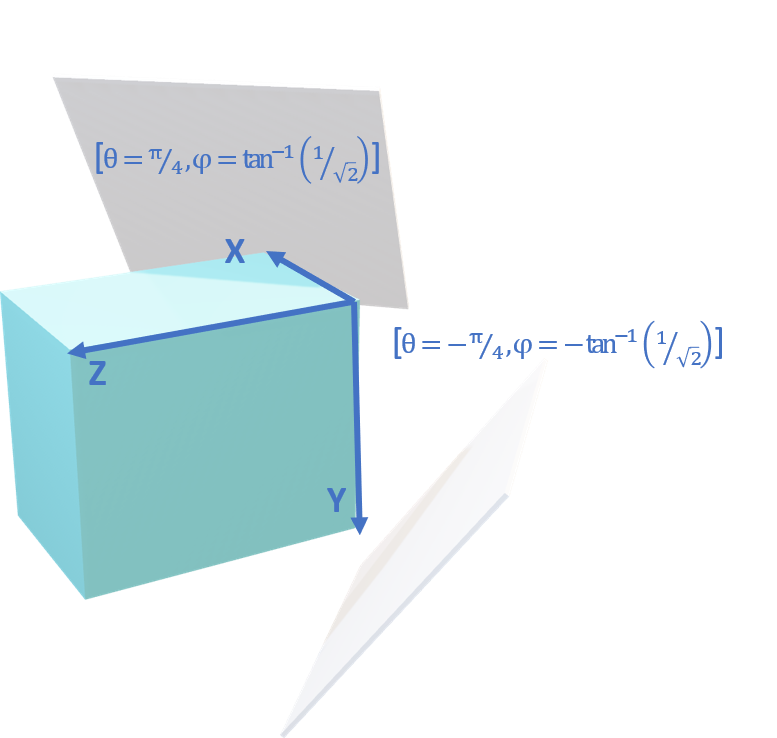}
\caption{Isometric projections}
\label{fig:isoprojects}
\end{subfigure}

\caption{A volume $V(x,y,z)$ and various orthographic projections}
\label{fig:projections}
\end{figure}

Given a volume (Figure \ref{fig:base}) of size $L\!\cdot\! M\!\cdot\! N$ and if we choose $(\theta\!=\!0,\phi\!=\!0)$ then the volume is projected along the $z$-axis onto the $(x,y)$ plane, resulting in a 2D integral image of size $L\!\cdot\! N$ (Figure \ref{fig:0projects}):
\begin{equation} \label{Ixy}
    I(i,j,0,0) = \sum_{\begin{smallmatrix}i=x\\j=y\end{smallmatrix}} V(x,y,z)
\end{equation}

Another convenient plane arises from $(\theta\!=\!0,\phi\!=\!\frac{\pi}{2})$, where the volume is projected along the $x$-axis onto the $(y,z)$ plane, giving a 2D integral image of size $M\!\cdot\! N$ (Figure \ref{fig:0projects}):
\begin{equation} \label{Iyz}
    I(i,j,0,{\textstyle\frac{\pi}{2}}) = \sum_{\begin{smallmatrix}i=y\\j=z\end{smallmatrix}} V(x,y,z)
\end{equation}

The third axis-aligned plane uses $(\theta\!=\!\frac{\pi}{2},\phi\!=\!0)$ and projects the volume along the $y$-axis onto the $(z,x)$ plane, resulting in a 2D integral image of size $N\!\cdot\! L$ (Figure \ref{fig:0projects}):
\begin{equation} \label{Izx}
    I(i,j,{\textstyle\frac{\pi}{2}},0) = \sum_{\begin{smallmatrix}i=z\\j=x\end{smallmatrix}} V(x,y,z)
\end{equation}

We can further construct discrete projections from aligned axonometric perspectives - for example using $(\theta\!=\!\frac{\pi}{4},\phi\!=\!0)$ results in a 2D integral image of size $(L\!+\!N\!-\!1)\!\cdot\! M$ (Figure \ref{fig:45projects}):
\begin{equation} \label{Ixzy}
    I(i,j,{\textstyle\frac{\pi}{4}},0) = \sum_{\begin{smallmatrix}i=x+z\\j=y\end{smallmatrix}} V(x,y,z)
\end{equation}

Similarly, rotating opposite using $(\theta\!=\!-\frac{\pi}{4},\phi\!=\!0)$ results in a 2D integral image of size $(L\!+\!N\!-\!1)\!\cdot\! M$ but over a different range and with negative indexing in the 2D image (Figure \ref{fig:45projects}):
\begin{equation} \label{Ix_zy}
    I(i,j,\!-{\textstyle\frac{\pi}{4}},0) = \sum_{\begin{smallmatrix}i=x-z\\j=y\end{smallmatrix}} V(x,y,z)
\end{equation}

Six such discrete projections can be generated by aligned rotations of $\pm\!\frac{\pi}{4}$ around each axis - equivalent to $\pm\!\arctan(1)$. Twelve may be generated from $\pm\!\arctan(\frac{1}{2})$ and $\pm\!\arctan(2)$, and so on. Similarly, discrete isometric projections result from rotations such as $(\theta\!=\!\frac{\pi}{4},\phi\!=\!\arctan(\frac{1}{\sqrt{2}}))$, the zero-padded image size is $(L\!+\!N\!-\!1)\!\cdot\! (L\!+\!2M\!+\!N\!-\!3)$ (Figure \ref{fig:isoprojects}):
\[
    I(i,j,{\textstyle\frac{\pi}{4}},{\textstyle\arctan(\frac{1}{\sqrt{2}}})) = \sum_{\begin{smallmatrix}i=x+z\\j=2y+x-z\end{smallmatrix}} V(x,y,z)
\]

\subsection{Moment Computation}
The 2D moments ($M_{pq}^{\theta:\phi}$) of the projection images can be computed and used to determine the 3D moments of the volumetric image. From equations \eqref{base_moments} and \eqref{Ixy}, the moments of the projection onto the plane $(x,y)$ are:
\begin{align}
    M_{pq}^{0:0} &= \sum_{i,j}I(i,j,0,0)i^{p}j^{q}\nonumber\\
    M_{pq}^{0:0} &= \sum_{i,j}(\sum_{z} V(i,j,z))i^{p}j^{q}\nonumber\\
    M_{pq}^{0:0} &= \sum_{i,j,z} V(i,j,z)i^{p}j^{q}z^{0}\nonumber\\
    \label{Mpq0}\therefore~ \mathcal{M}_{pq0} &= M_{pq}^{0:0} 
\end{align}

From equations \eqref{Iyz} and \eqref{Izx}, the 2D moments of projections onto the planes $(y,z)$ and $(z,x)$ are also used to determine some of the 3D moments:
\begin{align}
    \label{M0pq}\mathcal{M}_{0pq} &= M_{pq}^{0:\frac{\pi}{2}}\\
    \label{Mqop}\mathcal{M}_{q0p} &= M_{pq}^{\frac{\pi}{2}:0}
\end{align}

The 2D moments of these three projections allow the computation of most of the 3D moments. The exception is for moments where all axes $xyz$ are required. Up to the $4^{th}$ order these include: $\mathcal{M}_{111}$, $\mathcal{M}_{211}$, $\mathcal{M}_{121}$ and $\mathcal{M}_{112}$.
The derivation for $\mathcal{M}_{111}$ is as follows. Expanding the moment $M_{21}^{\frac{\pi}{4}:0}$ and using Equation \eqref{Ixzy}:
\begin{align}
    M_{21}^{\frac{\pi}{4}:0} &= \sum_{i,j}I(i,j,{\textstyle\frac{\pi}{4}},0)i^{2}j\nonumber\\
    M_{21}^{\frac{\pi}{4}:0} &= \sum_{i,j}(\sum_{\begin{smallmatrix}i=x+z\\j=y\end{smallmatrix}} V(x,y,z))i^{2}j\nonumber\\
    M_{21}^{\frac{\pi}{4}:0} &= \sum_{x,y,z} V(x,y,z)(x+z)^{2}y\nonumber\\
    M_{21}^{\frac{\pi}{4}:0} &= \sum_{x,y,z} V(x,y,z)(x^{2}y+2xyz+yz^{2})\nonumber\\
    M_{21}^{\frac{\pi}{4}:0} &= \mathcal{M}_{210}+2\mathcal{M}_{111}+\mathcal{M}_{012}\nonumber\\
    \label{M111}\therefore~ \mathcal{M}_{111} &= (M_{21}^{\frac{\pi}{4}:0}-\mathcal{M}_{210}-\mathcal{M}_{012})/2
\end{align}

Similarly, by using the $4^{th}$ order moments  $M^{\frac{\pi}{4}:0}_{22}$, $M^{-\frac{\pi}{4}:0}_{22}$, $M^{\frac{\pi}{4}:0}_{31}$, $M^{-\frac{\pi}{4}:0}_{31}$, etc, of the 2D projections, the following can be derived:
\begin{align}
    \label{M121}\mathcal{M}_{121} &= (M_{22}^{\frac{\pi}{4}:0}-\mathcal{M}_{220}-\mathcal{M}_{022})/2\\
    \label{M112}\mathcal{M}_{112} &= (M_{31}^{\frac{\pi}{4}:0}-M_{31}^{-\frac{\pi}{4}:0}-2\mathcal{M}_{013})/6\\
    \label{M211}\mathcal{M}_{211} &= (M_{31}^{\frac{\pi}{4}:0}+M_{31}^{-\frac{\pi}{4}:0}-2\mathcal{M}_{310})/6    
\end{align}

Using other discrete and orthographic integral projections along with their 2D moments, formulae for higher order 3D moments can be derived and their values computed efficiently. Furthermore, the computation of the 2D moments of the projections can be accomplished efficiently using a Discrete Radon Transform based algorithm \cite{diggin2020using}.

Given several 2D projections of the 3D volume will be used to compute the 3D moments, it is possible to invert the normal construction. Instead of performing a full tomographic sweep of the 3D artifact, several chosen 2D tomographic projections can be used instead of 3D reconstruction. This set of projections could be used to determine the 3D moments \cite{SALZMAN1990129} and therefore reduce the initial data acquisition demands and associated complexity.

\section{Example}
\begin{figure}[ht]
\centering
\includegraphics[width=1.0\linewidth]{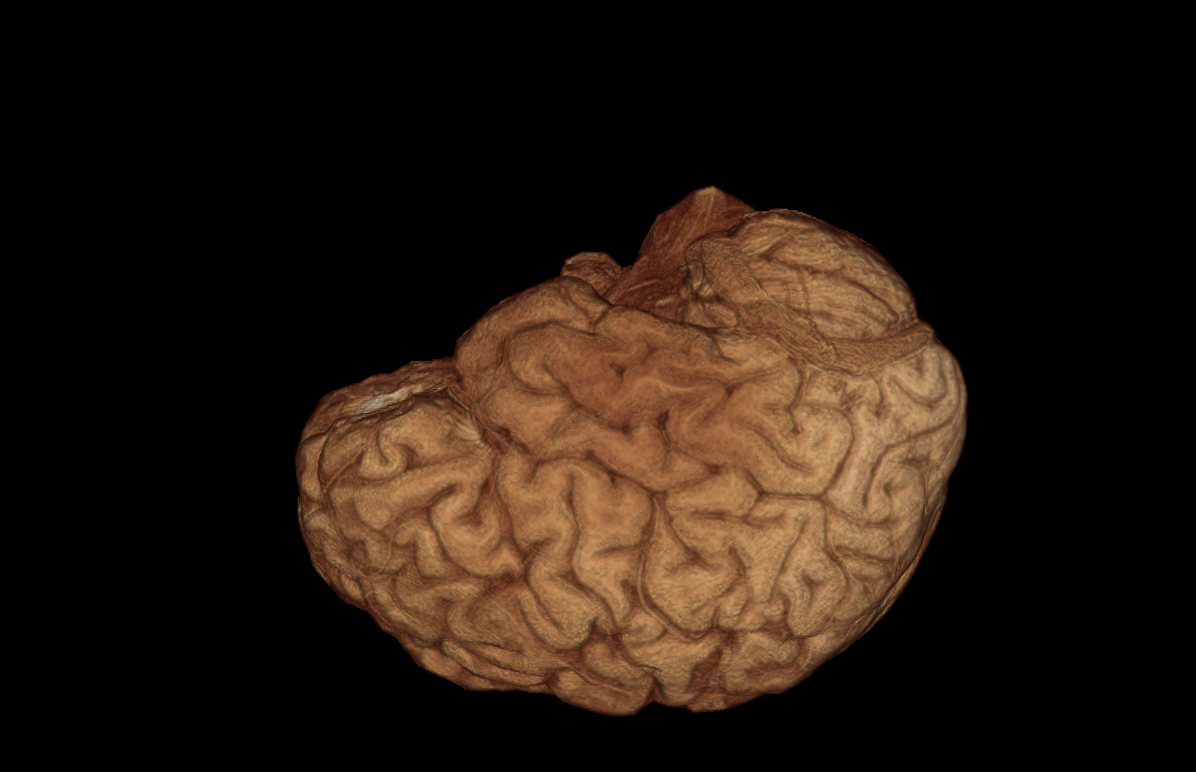}
\caption{Sample Brain MRI (color-enhanced)}
\label{fig:brainmri}
\end{figure}

To visualise an example of the DPM algorithm in practice, a volumetric image of a brain MRI was used, Figure \ref{fig:brainmri}. The object has been imaged with a 1mm spacing in the coronal and saggital directions and with a 0.4mm spacing in the axial direction. The image dimensions are 512x512x230 voxels. 

\begin{figure}[ht]
\begin{subfigure}{0.4\textwidth}
\includegraphics[width=1.0\linewidth]{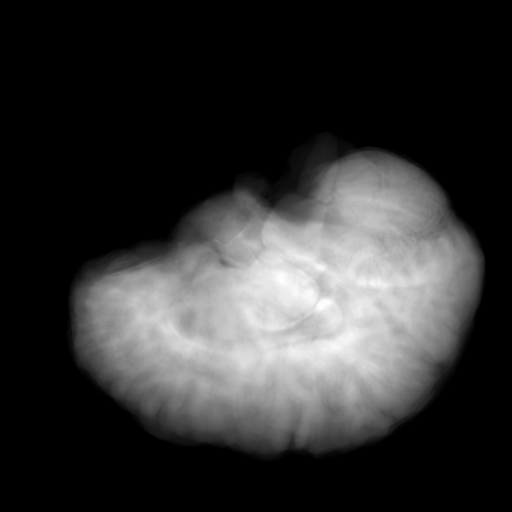}
\caption{2D projection image}
\label{fig:xy_plane1}
\end{subfigure}
\begin{subfigure}{0.5\textwidth}
\includegraphics[width=1.0\linewidth]{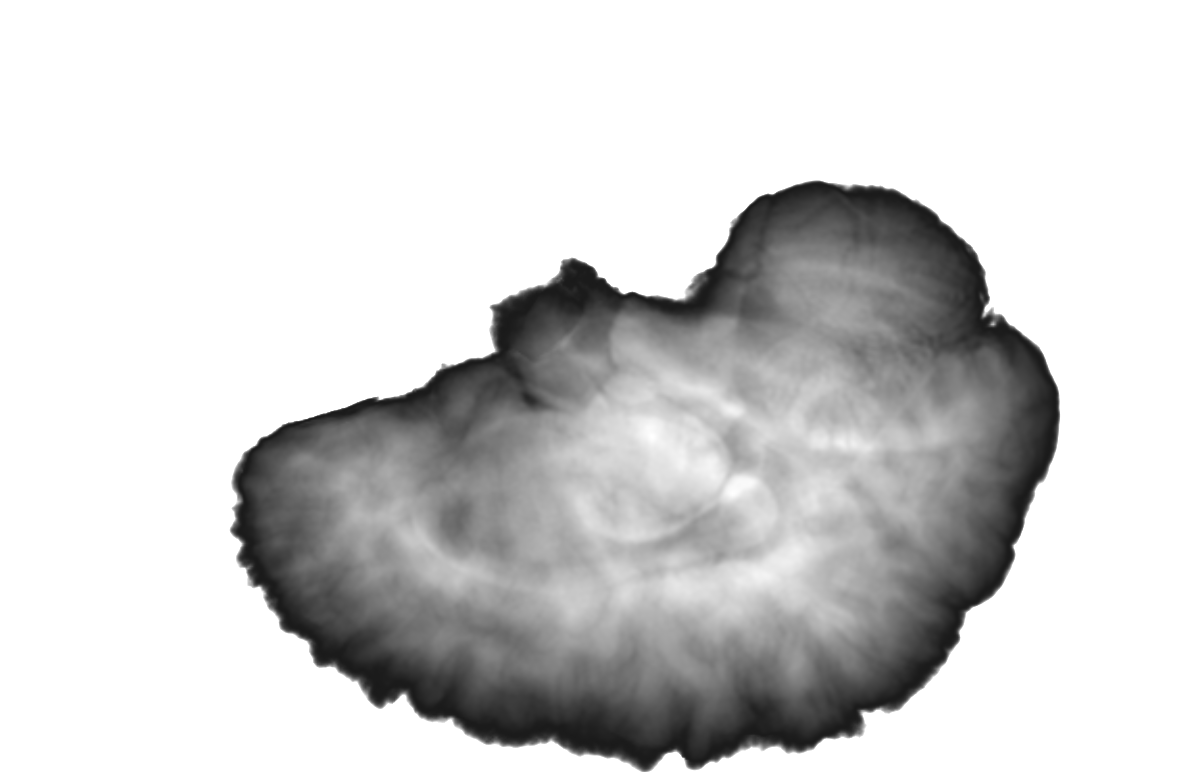}
\caption{3D front plane perspective}
\label{fig:xy_plane2}
\end{subfigure}
\begin{subfigure}{0.49\textwidth}
\includegraphics[width=1.0\linewidth]{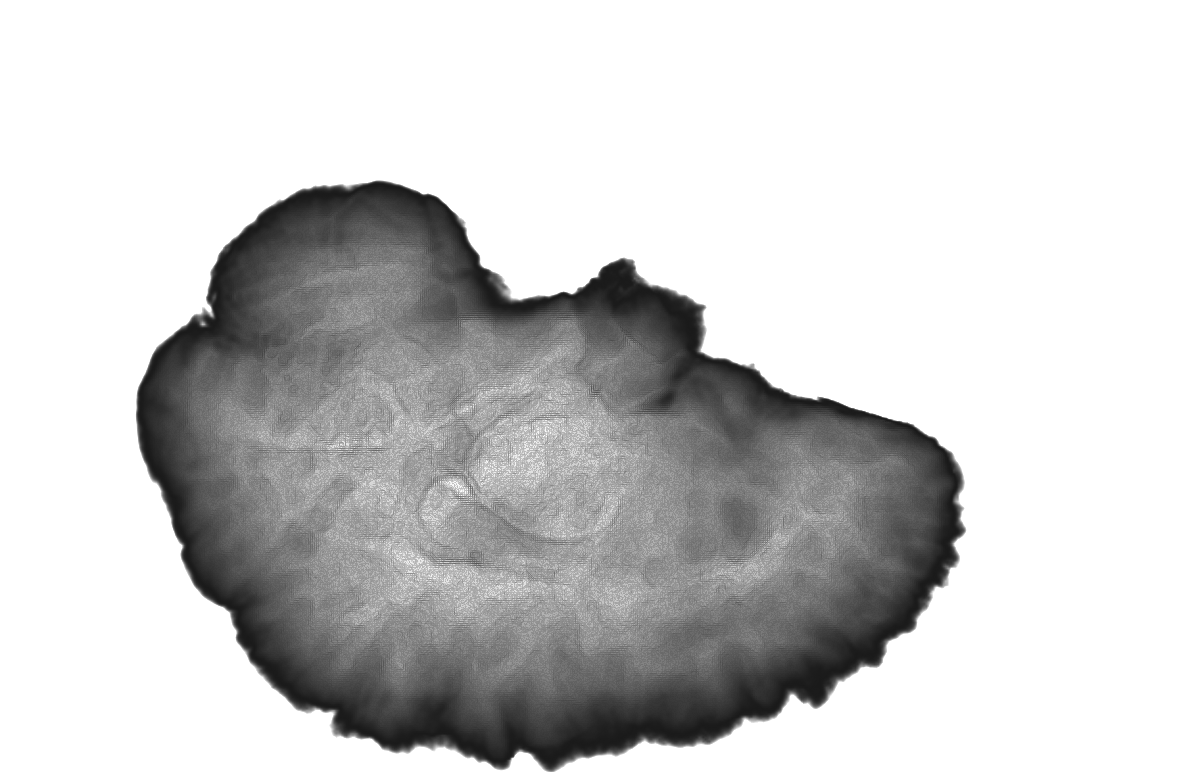}
\caption{3D rear plane perspective}
\label{fig:xy_plane3}
\end{subfigure}
\begin{subfigure}{0.5\textwidth}
\includegraphics[width=1.0\linewidth]{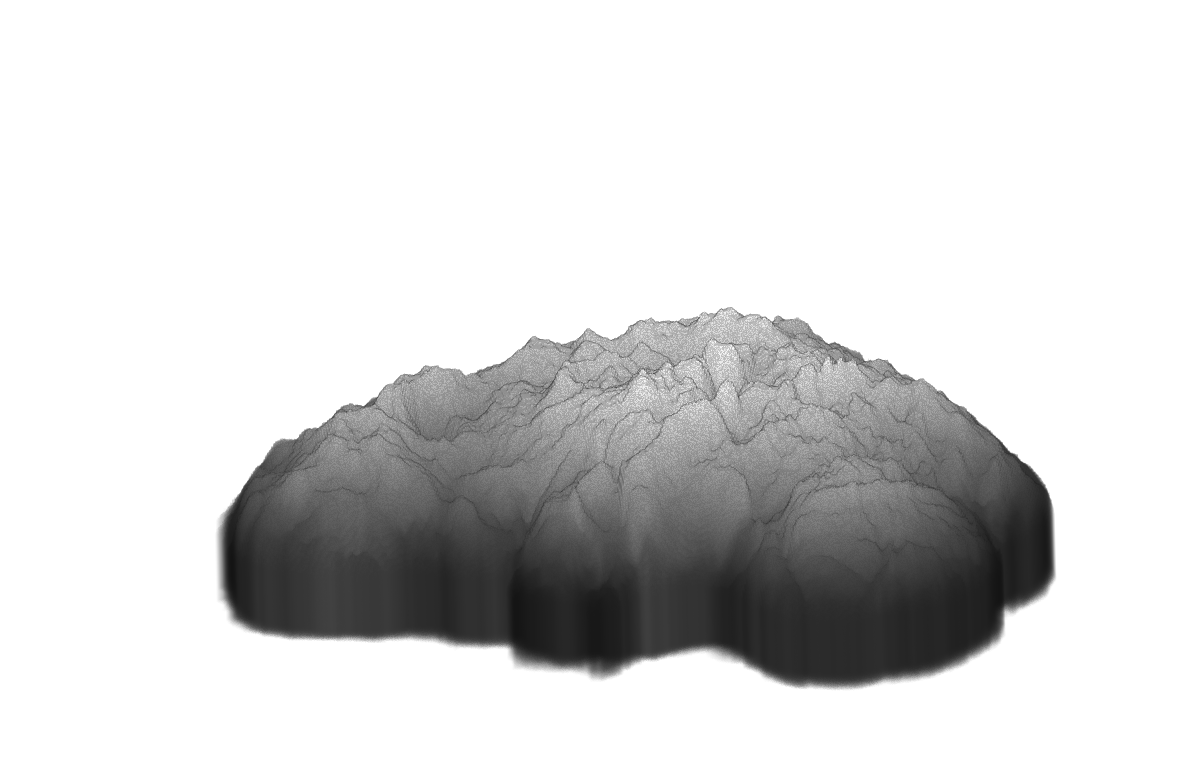}
\caption{3D isometric perspective}
\label{fig:xy_plane4}
\end{subfigure}
\caption{Projection onto $x,y$ plane}
\label{fig:xy_plane}
\end{figure}

\begin{figure}[ht]
\begin{subfigure}{0.49\textwidth}
\includegraphics[width=1.0\linewidth]{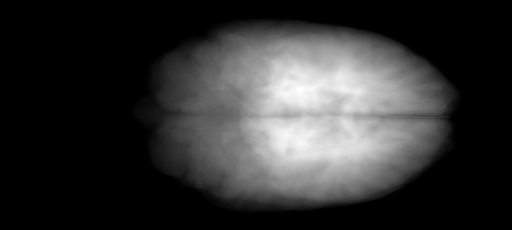}
\caption{}
\end{subfigure}
\begin{subfigure}{0.49\textwidth}
\includegraphics[width=1.0\linewidth]{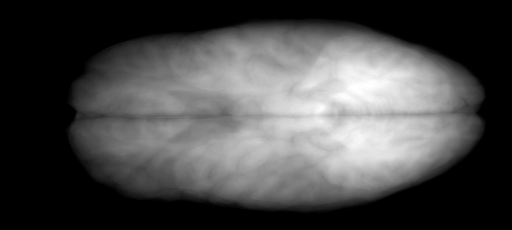}
\caption{}
\end{subfigure}
\caption{Projections onto (a) $y,z$ and (b) $z,x$ planes}
\label{fig:yz_zx_plane}
\end{figure}

\begin{figure}[ht]
\begin{subfigure}{0.49\textwidth}
\includegraphics[width=1.0\linewidth]{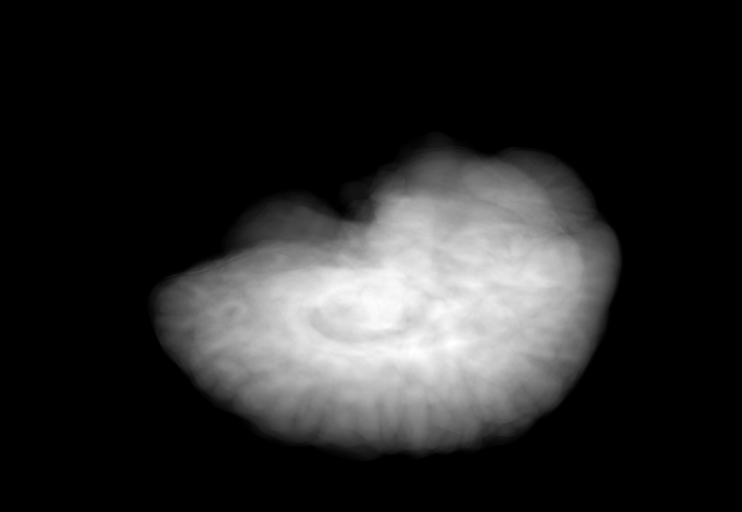}
\caption{}
\end{subfigure}
\begin{subfigure}{0.49\textwidth}
\includegraphics[width=1.0\linewidth]{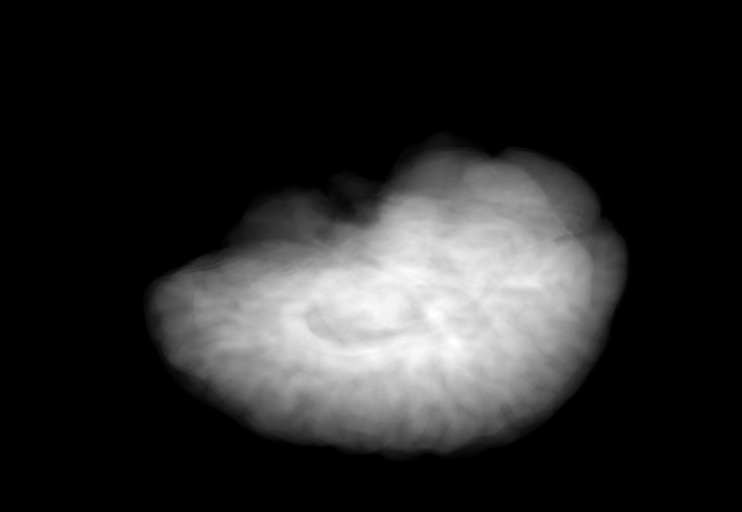}
\caption{}
\end{subfigure}
\caption{Projections onto (a) $x\!+\!z,y$ and (b) $x\!-\!z,y$ planes}
\label{fig:xz_y_plane}
\end{figure}

To compute the $4^{th}$ order 3D moments, we require five projections. We choose, for economies of computation, projections based on equations \eqref{Ixy}, \eqref{Iyz}, \eqref{Izx}, \eqref{Ixzy} and \eqref{Ix_zy}. This leads to five 2D images as presented in Figures \ref{fig:xy_plane}, \ref{fig:yz_zx_plane} and \ref{fig:xz_y_plane}. These 2D images are generated by indexing through the volume data and accumulating the voxel values into each image. Figures \ref{fig:xy_plane2}, \ref{fig:xy_plane3} and \ref{fig:xy_plane4} provide additional visualisation perspectives to the 2D projection onto the (x,y) plane. The accumulation is very efficient as it is performed with pure addition.

From these five 2D images, we now wish to compute their 2D moments. For this, we choose the efficient DRT algorithm \cite{diggin2020using} that further projects each of the 2D images to a series of 1D integrals. Again, this step is an efficient process as it too is performed with pure addition.
Based on that DRT algorithm, the moments of these 1D integrals are computed and then combined to generate all the $4^{th}$ order 2D moments for these five 2D images.

Finally, using equations \eqref{Mpq0}, \eqref{M0pq}, \eqref{Mqop}, \eqref{M111}, \eqref{M112}, \eqref{M121} and \eqref{M211}, we can compute all the $3^{rd}$ or $4^{th}$ order 3D moments of the Brain MRI. Total computation time is $\approx$20ms for $3^{rd}$ order and is $\approx$30ms for the $4^{th}$ order.

Computation of the object features, such as centroid, is now possible:
\begin{align*}
    \bar{x}&=\frac{\mathcal{M}_{100}}{\mathcal{M}_{000}} &  \bar{y}&=\frac{\mathcal{M}_{010}}{\mathcal{M}_{000}} &  \bar{z}&=\frac{\mathcal{M}_{001}}{\mathcal{M}_{000}}
\end{align*}
From these, the central moments, $\mu_{pqr}$, may be computed:
\[
    \mu_{pqr}= \sum V(x,y,z)(x\!-\!\bar{x})^{p}(y\!-\!\bar{y})^{q}(z\!-\!\bar{z})^{r}
\]
The central moments are invariant to translation. The central moments can then be normalised for scale invariance by:
\begin{equation}
    \eta_{pqr}=\frac{\mu_{pqr}}{\mu_{000}^{\frac{p+q+r}{3}+1}}
\end{equation}
However, in most tomography-based imaging systems, the physical spacing is programmable. In an MRI DiCOM sequence, axial imaging is constituted with a series of, say, 512x512 images. In this case, each image element may have a physical spacing of 1.0mm in the $x-,y$-axes and each slice in the axial plane (or $z$-axis) may have a physical spacing of 0.4mm. This is representative of our example. This means that the computed moments from the DPM algorithm are only valid for the virtual voxel spacing. To compensate for the scale mapping from virtual to physical, we can normalise the computed central moments as follows:
\[
    \mu^{'}_{pqr} = \alpha^{1+p}\beta^{1+q}\delta^{1+r}\mu_{pqr}
\]
The quantities, $\alpha$, $\beta$ and $\delta$ are the scale factors in each image axis.

Given the moments are now available to us relative to the physical coordinate frame, we can compute several other metrics, such as orientation, elongation (length/width ratios), sphere of gyration, as well as feature descriptors that are invariant to translation, rotation and scale \cite{SadjadiArticle}, \cite{flusser_tensor}.

An interesting avenue for further investigation revolves around performing pre-processing in projection or integral space. For example, rather than conducting background segmentation or subtraction on the 3D volume, we can perform that operation on the projection images or the 1D integrals. Obviously there are resulting savings in computational complexity and processing time.

\section{Implementation Details} \label{implementation}

To take advantage of the computational efficiencies of the DPM algorithm, we outline some of the important implementation details to assist in construction. The pseudo-code to generate the projections is given in Listing 1. This shows the mechanism by which the 2D projection images are constructed. It is clear that pure discrete addition of voxels to pixels occurs.

\begin{lstlisting}[language=C++, caption=Projection Pseudo-Code]

    // Volume = 3D image volume
    // XY, YZ, ZX, X_plusZY
    // and X_minus_ZY = 2D projection images
    For k = 1 to Depth
        For j = 1 to Height
    
            pXZY  = address_of( X_plus_ZY[k,j] )
            pX_ZY = address_of( X_minus_ZY[Depth - k - 1, j] )
    
            For i = 1 to Width
                XY[i, j] += Volume[i, j, k]
                YZ[j, k] += Volume[i, j, k]
                ZX[k, i] += Volume[i, j, k]
                pXZY[i]  += Volume[i, j, k]
                pX_ZY[i] += Volume[i, j ,k]
            EndFor
        EndFor
    EndFor
    
\end{lstlisting}

The DRT algorithm to compute the 2D moments projects each of the 2D images computed in Listing 1 onto five 1D integrals as illustrated in Listing 2. The accumulation into the integrals is a pure discrete additive process. Therefore, the first two steps in the computation chain are straightforward and trivial to construct.

\begin{lstlisting}[language=C++, caption=DRT Integral Pseudo-Code]

    // Image = 2D image - Width x Height
    // Horiz = integral onto the Y axis
    // Vert = integral onto the X axis
    // Diagonal = integral onto 45 degree axis
    // AntiDiagonal = integral onto 135 degree axis
    // X2Y = integral onto atan(0.5) degree axis
    
    pDiag = address_of( Diagonal[0] )
    pAnti = address_of( AntiDiagonal[Height - 1] )
    pX2Y  = address_of( X2Y[0] )

    For y = 1 to Height
        For x = 1 to Width
            Vert[x]  += Image[x, y]
            Horiz[y] += Image[x, y]
            pDiag[x] += Image[x, y]
            pAnti[x] += Image[x, y]
            pX2Y[x]  += Image[x, y]
        EndFor

        pX2Y += 2
        pDiag++
        pAnti--
        
    EndFor
    
\end{lstlisting}

By iterating linearly through the 3D volume data, the 2D image data and the 1D integral data, the algorithm capitalises on processor caching and pipelining optimisation and so captures significant time advantages.

For $4^{th}$ order, the moments of each of the five 2D images are computed according to the DRT algorithm \cite{diggin2020efficient}. For each image, the five 1D integrals are used to compute their equivalent 1D moments. These are then combined to provide the 2D moments for each image.

The 2D moments, $M^{-\frac{\pi}{4}:0}_{pq}$, from the projection in Equation \eqref{Ix_zy} need to be adjusted due to the requirement of negative indexing. This simply means an offset so that the moment origin is positioned at $\hat{x}=N\!-\!1$ in the 2D image. This can be accomplished for the relevant moments as follows:
\begin{align*}
    M^{-\frac{\pi}{4}:0}_{31}&=M^{-\frac{\pi}{4}:0}_{31}-3M^{-\frac{\pi}{4}:0}_{21}\hat{x}+3M^{-\frac{\pi}{4}:0}_{11}\hat{x}^{2}-M^{-\frac{\pi}{4}:0}_{01}\hat{x}^{3}\\
    M^{-\frac{\pi}{4}:0}_{13}&=M^{-\frac{\pi}{4}:0}_{13}-M^{-\frac{\pi}{4}:0}_{03}\hat{x}
\end{align*}

Finally, use equations \eqref{Mpq0}, \eqref{M0pq}, \eqref{Mqop}, \eqref{M111}, \eqref{M112}, \eqref{M121} and \eqref{M211} to compute all the relevant 3D moments.

\section{Results} \label{results}
We implemented the naïve computational approach, based on Equation \eqref{3d_moments}, to provide a reference for computational cost. In addition, as the OpenCV Library is the only other available source of reference to efficiently compute grayscale moments, we extended its 2D OpenCV moment implementation. The base algorithm used in the OpenCV library was extended to compute the $3^{rd}$ and $4^{th}$ order 3D moments. The naïve and OpenCV algorithms were measured and compared against the DPM algorithm detailed in this paper. All the algorithms have been coded, compiled and optimised for the test processor instruction set. Then, using a large sample size to maximise processor availability, we selected the fastest exemplar for each algorithm and each test candidate. The comparison of execution times was gathered for differing volumetric image sizes, measured in MegaBytes (MB). All algorithms were executed on an Intel Core i5 $7^{th}$ Generation CPU. 
\begin{table}[h]
\begin{center}
\begin{tabular}{ | r | rr|SS|SS| }
\hline
\textbf{Size} & \multicolumn{2}{ c | }{\textbf{Naïve (ms)}} & \multicolumn{2}{ c | }{\textbf{OpenCV (ms)}} & \multicolumn{2}{ c | }{\textbf{DPM (ms)}}\\
(MB) & $3^{rd}$ & $4^{th}$ & \multicolumn{1}{ c }{$3^{rd}$} & \multicolumn{1}{ c |}{$4^{th}$} & \multicolumn{1}{ c }{$3^{rd}$} & \multicolumn{1}{ c |}{$4^{th}$}\\
\hline\hline
1 & 11 & 23 & 1.66 & 2.00 & 0.30 & 0.44\\
\hline
2 & 23 & 46 & 3.36 & 4 & 0.56 & 0.90\\
\hline
4 & 47 & 94 & 6.70 & 8.07 & 1.00 & 1.46\\
\hline
8 & 95 & 188 & 13.6 & 16.75 & 2.00 & 2.75\\
\hline
16 & 192 & 377 & 28.0 & 34.1 & 4.03 & 5.3\\
\hline
32 & 386 & 760 & 57.0 & 68.95 & 8.25 & 10.95\\
\hline
64 & 772 & 1447 & 115.9 & 130.4 & 17.60 & 21.23\\
\hline
128 & 1466 & 2885 & 221.5 & 262.0 & 37.5 & 41.0\\
\hline
256 & 2927 & 5793 & 445.0 & 525.2 &63.5 & 84.6\\
\hline
512 & 5905 & 12200 & 892.1 & 1052.4 & 130.4 & 173.8\\
\hline
1024 & 12772 & 24646 & 1933.0 & 2198.0 & 273.5 & 400.1\\
\hline
\end{tabular}
\caption{Comparative Computational Times vs Volumetric Image Sizes for Naïve, OpenCV and DPM for $3^{rd}$ and $4^{th}$ order 3D moments.}
\label{table:1}
\end{center}
\end{table}
\begin{figure}[ht]
\centering
\includegraphics[scale=0.65]{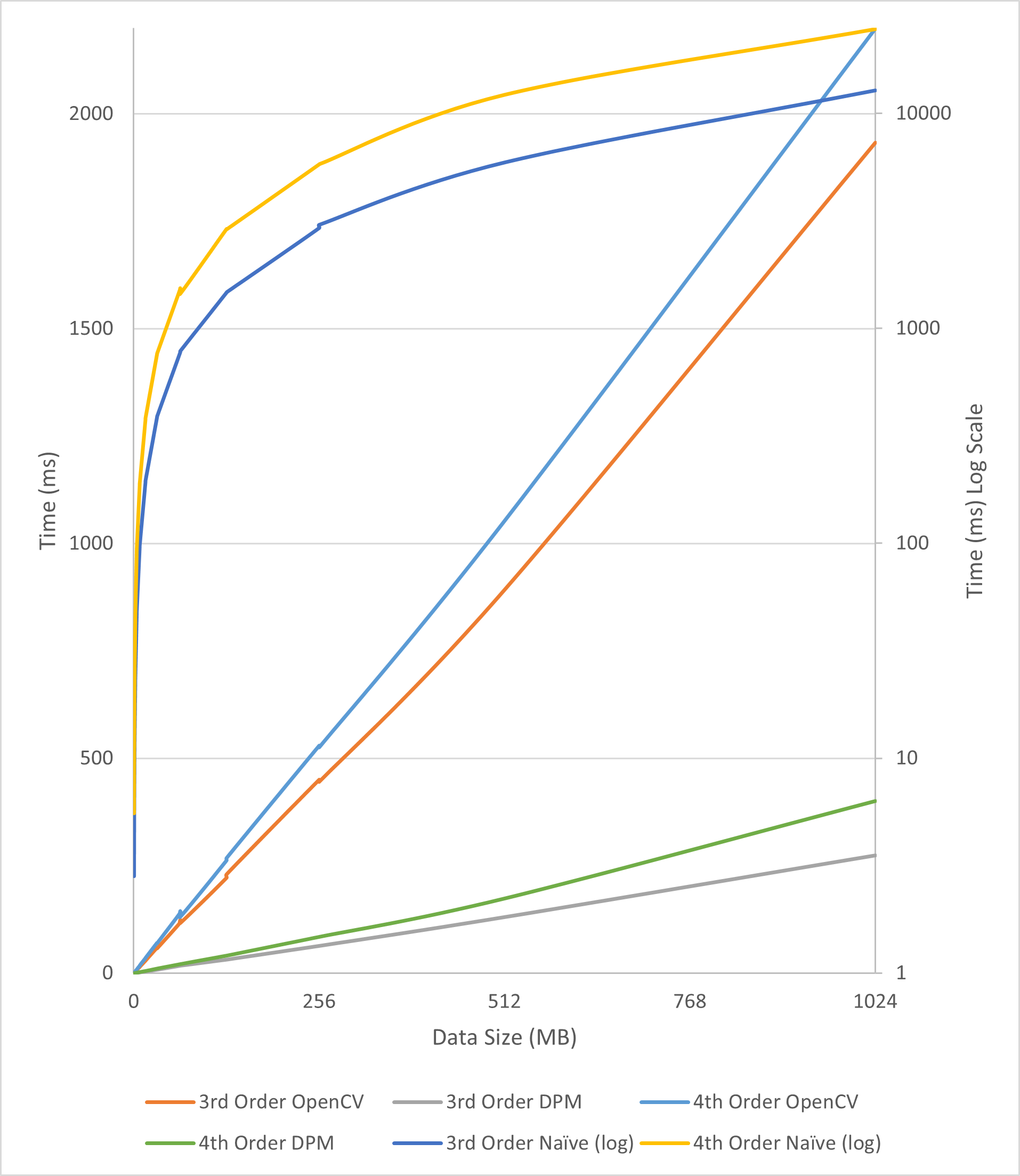}
\caption{Comparative Computational Cost Vs Image Sizes for Naïve, OpenCV and DPM.}
\label{fig:plot3}
\end{figure}

Table \ref{table:1} and Figure \ref{fig:plot3} show the measurements and plot comparing the time needed to compute the 3D moments of the $3^{rd}$ and $4^{th}$ order for a given volume size. The smallest volume used was a 128x128x64 voxel image, with the largest at 1024x1024x1024 voxels. The y-axis illustrates the execution time in milliseconds. The secondary y-axis is also in milliseconds but in $\log_{10}$ for easier plotting. The x-axis is the volume data size in MegaBytes.

It is evident that the naïve approach does not scale well. The cost to execute the $4^{th}$ order moments using the naïve approach is twice the cost for the $3^{rd}$ order. It requires 12 and 24 seconds to compute the $3^{rd}$ and $4^{th}$ order moments for a 1GB volume.

The OpenCV algorithm performs significantly better, circa 6x to 10x better than the naïve approach. Is has only a $\approx$20\% overhead when computing the $4^{th}$ order versus the $3^{rd}$ order. It requires $\approx$1.9 and $\approx$2.2 seconds to compute the $3^{rd}$ and $4^{th}$ order moments for a 1GB volume.

The DPM algorithm performs best, 5x to 10x better than the OpenCV approach. It has a $\approx$30\% overhead when computing the $4^{th}$ order versus the $3^{rd}$ order. The DPM algorithm needed only $\approx$0.3 and 0.4 seconds to compute the $3^{rd}$ and $4^{th}$ order moments for a 1GB volume.

Note that all algorithms exhibit an approximate doubling of processing time for an equivalent doubling in data size.

\section{Analysis} \label{analysis}
To simplify the analysis, consider an $n\times n\times n$ volume. The naïve implementation to compute all 3D moments up to the $4^{th}$ order using equation \eqref{3d_moments} requires $37n^{3}$ multiplication and $35n^{3}$ additions. Extending the OpenCV implementation from 2D to 3D, the same moment computation requires $4n^{3}+14n^{2}+20n$ multiplications with $5n^{3}+20n^{2}+20n$ additions. 

As summarised in Table \ref{table:2}, for the DPM algorithm, the generation of the 2D images from the 3D volume is computationally trivial - requiring $4n^{3}$ additions for $3^{rd}$ order and $5n^{3}$ additions for $4^{th}$ order. No multiplications are required for this step.

For the second step, the construction of the 1D integrals of the 2D images is also computationally trivial. For all 2D images, it requires $28n^{2}$ additions for $3^{rd}$ order and $35n^{2}$ additions for $4^{th}$ order. Again, no multiplications are involved.

The computation of the moments from the 1D integrals requires $54n$ multiplications for $3^{rd}$ order and $126n$ multiplications for $4^{th}$ order. The computation needs $60n$ additions for $3^{rd}$ order and $132n$ additions for $4^{th}$ order. Following these steps is the computation of the 2D moments from the 1D integral moments and then the 3D moments from the 2D moments. This is a constant computation and is straightforward.

\begin{table}[h]
\begin{center}
\begin{tabular}{ | l | c | c | c | }
\hline
\textbf{Method} & \textbf{+/$\times$} & \textbf{$3^{rd}$ Order} & \textbf{$4^{th}$ Order}\\
\hline\hline
Naïve & $+$ & $20n^{3}$ & $35n^{3}$\\
\hline
Naïve & $\times$ & $21n^{3}$ & $37n^{3}$\\
\hline
\hline
OpenCV & $+$ & $4n^{3}+10n^{2}+20n$ & $5n^{3}+20n^{2}+20n$\\
\hline
OpenCV & $\times$ & $3n^{3}+7n^{2}+11n$ & $4n^{3}+14n^{2}+20n$\\
\hline
\hline
DPM & $+$ & $4n^{3}+28n^{2}+60n$ & $5n^{3}+35n^{2}+132n$ \\
\hline
DPM & $\times$ & $54n$ & $126n$\\
\hline
\end{tabular}
\caption{Scale of multiplications and additions of the Naïve, OpenCV and DPM for $3^{rd}$ and $4^{th}$ order 3D moments.}
\label{table:2}
\end{center}
\end{table}

Therefore for a volume of size $400\times400\times400$, naïve vs OpenCV vs DPM have the following multiplication ratios for $3^{rd}$ order moments approximately: 62,222 : 8,888 : 1. The multiplication ratios for $4^{th}$ order moments approximately are 188,000 : 5,000 : 1. The computation for $3^{rd}$ and $4^{th}$ order addition is $4n^{3}$ and $5n^{3}$ for the OpenCV method and the DPM algorithm - although the naïve method requires $\approx5\times$ and $\approx7\times$ relative to the others.

The naïve and OpenCV approaches both exhibit $\mathcal{O}(n^{3})$ multiplicative complexity whereas the DPM algorithm is $\mathcal{O}(n)$. All methods are $\mathcal{O}(n^{3})$ additive complexity. In practical terms, each algorithm has $\mathcal{O}(n^{3})$ complexity to load data for processing and store the results after computation. The DPM algorithm reduces the total number of operations by eliminating much of the per voxel multiplication. This is the main reason for its superior performance.

\section{Conclusion} \label{conclusion}
The DPM algorithm provides significant computational advantages as volume size increases. It is shown to be $50\!-\!60\times$ faster than the direct naïve computation and $5\!-\!7\times$ faster than an optimal OpenCV algorithm. The DPM algorithm performs well for both binary or grey-level volumetric images and does not require any surface tracking or morphological pre-processing. The approach demonstrates an efficient means to generate $3^{rd}$ and $4^{th}$ order moments of 3D artifacts. By generating other discrete projections we can extend the algorithm to higher order moments, while retaining computational efficiency. The resultant moment invariants have uses as feature descriptors, ideally as input to classifiers or neural inference engines.

Further work to extend the approach to higher order moments in a formal and general fashion is warranted. So too is extending the algorithmic approach to multi-dimensional data-sets, e.g. temporal related signals. More simply, there are opportunities to capitalise on other transformations to the data in projection or integral space and then compute the resulting moments. This is instead of conducting those transformations in higher-dimensional space and suffering the resulting computational cost.

\section{Acknowledgements}
The authors wish to acknowledge the volume datasets provided by the Naval Postgraduate School at \url{https://savage.nps.edu/Savage/VolumeRendering/datasets/nrrd/}.
The authors also acknowledge the datasets provided via \url{https://slicer.org} - contributors: Steve Pieper (Isomics), Benjamin Long (Kitware) and Jean-Christophe Fillion-Robin (Kitware).
\printbibliography

@article{Sossaarticle,
author = {Sossa-Azuel   , Humberto and Cuevas, Francisco and Aguilar-Ibáñez, Carlos and Benitez-Muñoz, Héctor},
year = {2007},
month = {06},
pages = {111-123},
title = {3-D Cartesian Geometric Moment Computation using Morphological Operations and its Application to Object Classification},
volume = {8},
journal = {Ingeniería, investigación y tecnología},
doi = {10.22201/fi.25940732e.2007.08n2.010}
}

@misc{patent:7274810,
 title     = {System and method for three-dimensional image rendering and analysis},
 number    = {7274810},
 author    = {Reeves, Anthony P. and Kostis, William J. and Henschke, Claudia and Yankelevitz, David},
 year      = {2007},
 month     = {September},
 url       = {https://www.freepatentsonline.com/7274810.html}
}

@inproceedings{flusser_tensor,
author = {Suk, Tomás and Flusser, Jan},
year = {2011},
month = {08},
pages = {212-219},
title = {Tensor Method for Constructing 3D Moment Invariants},
doi = {10.1007/978-3-642-23678-5_24}
}

@article{brain_morohometry,
author = {Mangin, Jean-François and Poupon, Fabrice and Duchesnay, Edouard and Rivière, Denis and Cachia, Arnaud and Collins, Louis and Evans, Alan and Régis, Jean},
year = {2004},
month = {09},
pages = {187-196},
title = {Brain morphometry using 3D moment invariants},
volume = {8},
journal = {Medical Image Analysis},
doi = {10.1016/j.media.2004.06.016}
}

@book{deans2013radon,
  title={The Radon Transform and Some of Its Applications},
  author={Deans, S.R.},
  isbn={9780486788548},
  series={Dover Books on Mathematics Series},
  url={https://books.google.ie/books?id=BXiXswEACAAJ},
  year={2013},
  publisher={Dover Publications, Incorporated}
}

@article{hammer,
author = {Shen, Dinggang and Davatzikos, Christos},
year = {2002},
month = {12},
pages = {1421-39},
title = {HAMMER: Hierarchical attribute matching mechanism for elastic registration},
volume = {21},
journal = {IEEE transactions on medical imaging},
doi = {10.1109/TMI.2002.803111}
}

@inproceedings{brain_segments,
author = {Jabarouti Moghaddam, Mostafa and Soltanian-Zadeh, Hamid},
year = {2009},
month = {02},
pages = {326-37},
title = {Automatic Segmentation of Brain Structures Using Geometric Moment Invariants and Artificial Neural Networks},
volume = {21},
isbn = {978-3-642-02497-9},
journal = {Information processing in medical imaging : proceedings of the ... conference},
doi = {10.1007/978-3-642-02498-6_27}
}

@article{artery_moments,
author = {Chen, Kun and Zhang, Y. and Pohl, Kilian and Syeda-Mahmood, Tanveer and Song, Zhihuan and Wong, Stephen},
year = {2010},
month = {08},
pages = {3133-7},
title = {Coronary Artery Segmentation Using Geometric Moments based Tracking and Snake-Driven Refinement},
volume = {2010},
journal = {Conference proceedings : ... Annual International Conference of the IEEE Engineering in Medicine and Biology Society. IEEE Engineering in Medicine and Biology Society. Conference},
doi = {10.1109/IEMBS.2010.5627192}
}

@article{blood_moments,
author = {Reuzé, Patrick and Coatrieux, Jean-Louis and Luo, Limin and Dillenseger, Jean-Louis},
year = {1993},
month = {11},
pages = {},
title = {A 3-D moment based approach for blood vessel detection and quantification in MRA},
volume = {1},
journal = {Technology and health care : official journal of the European Society for Engineering and Medicine},
doi = {10.3233/THC-1993-1209}
}

@inproceedings{fmri_moments,
author = {Ng, Bernard and Abugharbieh, R. and Huang, Xuemei and McKeown, Martin},
year = {2006},
month = {07},
pages = {63- 63},
title = {Characterizing fMRI Activations within Regions of Interest (ROIs) Using 3D Moment Invariants},
volume = {2006},
isbn = {0-7695-2646-2},
journal = {Proceedings of the IEEE Computer Society Conference on Computer Vision and Pattern Recognition},
doi = {10.1109/CVPRW.2006.52}
}

@inproceedings{Yang,
author = {Yang, Luren and Albregtsen, Fritz and Taxt, Torfinn},
year = {1995},
month = {01},
pages = {649-654},
title = {Fast Computation of 3-D Geometric Moments Using a Discrete Gauss' Theorem.}
}

@INPROCEEDINGS{Li_Ma,  
author={ {BingCheng Li} and  {Song De Ma}},  booktitle={Proceedings of 12th International Conference on Pattern Recognition},   
title={Efficient computation of 3D moments},   year={1994},  volume={1},  number={},  pages={22-26 vol.1},  doi={10.1109/ICPR.1994.576218}
}

@article{SALZMAN1990129,
title = "A method of general moments for orienting 2D projections of unknown 3D objects",
journal = "Computer Vision, Graphics, and Image Processing",
volume = "50",
number = "2",
pages = "129 - 156",
year = "1990",
issn = "0734-189X",
doi = "https://doi.org/10.1016/0734-189X(90)90038-W",
url = "http://www.sciencedirect.com/science/article/pii/0734189X9090038W",
author = "David B Salzman"
}

@article{cadmoments,
author = {Cybenko, George and Bhasin, Aditya and Cohen, Kurt},
year = {1999},
month = {09},
pages = {},
title = {Pattern Recognition of 3D CAD Objects : Towards an Electronic Yellow Pages of Mechanical Parts},
volume = {1},
journal = {Smart Engineering Systems Design}
}

@article{yang2,
author = {Yang, Luren and Albregtsen, Fritz and Taxt, Torfinn},
year = {1997},
month = {03},
pages = {97-108},
title = {Fast Computation of Three-Dimensional Geometric Moments Using a Discrete Divergence Theorem and a Generalization to Higher Dimensions},
volume = {59},
journal = {CVGIP: Graphical Model and Image Processing},
doi = {10.1006/gmip.1997.0418}
}

@inproceedings{Berjon,
       booktitle = {Proceedings of the SPIE. Parallel Processing for Imaging Applications},
          volume = {7872},
           title = {A parallel implementation of 3D Zernike moment analysis},
          author = {Daniel Berj{\'o}n D{\'i}ez and Sergio Arnaldo Duart and Francisco Mor{\'a}n Burgos},
       publisher = {SPIE},
            year = {2011},
             url = {http://oa.upm.es/12233/}
}

@misc{diggin2020efficient,
      title={Efficient Computation of Higher Order 2D Image Moments using the Discrete Radon Transform}, 
      author={William Diggin and Michael Diggin},
      year={2020},
      eprint={2009.09898},
      archivePrefix={arXiv},
      primaryClass={cs.CV}
}

@article{HuArticle,
  author="MK Hu",
  journal={IRE Transactions on Information Theory}, 
  title={Visual pattern recognition by moment invariants}, 
  year={1962},
  volume={8},
  number={2},
  pages={179-187}
}

@article{SadjadiArticle,
author = {Sadjadi, Firooz and Hall, Ernest},
year = {1980},
month = {04},
pages = {127 - 136},
title = {Three-Dimensional Moment Invariants},
volume = {2},
journal = {Pattern Analysis and Machine Intelligence, IEEE Transactions on},
doi = {10.1109/TPAMI.1980.4766990}
}

@article{opencv_library,
    author="G Bradski",
    journal = {Dr. Dobb's Journal of Software Tools},
    title = {{The OpenCV Library}},
    year = {2000}
}

@article{systolic_moments,
author = {Liu, Jianguo and Chen, Shaobo and Wang, Hai},
year = {2009},
month = {10},
pages = {},
title = {Global systolic array for fast computation of 3D moment invariants},
journal = {Proc SPIE},
doi = {10.1117/12.833710}
}

@misc{diggin2020using,
    title={Using the discrete radon transformation for grayscale image moments},
    author="W Diggin and M Diggin",
    year={2020},
    eprint={2008.11083},
    archivePrefix={arXiv},
    primaryClass={cs.CV}
}

\end{document}